\documentclass[letterpaper, 10 pt, conference]{ieeeconf}  

\IEEEoverridecommandlockouts                              

\usepackage{graphicx} 

\title{\LARGE \bf
Learning to Catch Piglets in Flight
}

\author{Ozan \c{C}atal, Lawrence De Mol, Tim Verbelen and Bart Dhoedt$^{*}$
\thanks{\{firstname\}.\{lastname\}@ugent.be\newline}%
\thanks{$^{*}$ All authors are associated with Ghent University}%
}

\begin{document}

\maketitle
\thispagestyle{empty}
\pagestyle{empty}

\begin{abstract}
Catching objects in-flight is an outstanding challenge in robotics. In this paper, we present a closed-loop control system fusing data from two sensor modalities: an RGB-D camera and a radar. To develop and test our method, we start with an easy to identify object: a stuffed Piglet. We implement and compare two approaches to detect and track the object, and to predict the interception point. A baseline model uses colour filtering for locating the thrown object in the environment, while the interception point is predicted using a least squares regression over the physical ballistic trajectory equations. A deep learning based method uses artificial neural networks for both object detection and interception point prediction. We show that we are able to successfully catch Piglet in 80\% of the cases with our deep learning approach.
\end{abstract}

\section{INTRODUCTION}

Current robot hardware exists to provide very fine grained and fast joint control, as well as accurate force-torque sensors. This robot hardware, in combination with an advanced perception system, enables robots to complete more and more complex tasks, while executing these tasks at a much higher speed than humans could achieve. Yet, there exist tasks that humans and even toddlers can do with ease, but that are very difficult for robots, for example catching a ball.
In this paper we propose a system to address the challenge of catching an object in flight with a robotic manipulator. This problem requires the combination of accurately measuring the current location of the flying object, predicting the trajectory that the object will follow and moving the robotic arm to the interception position. To complete this task successfully, these actions need to happen within the short flying time of the object.

To detect and track an object we fuse information coming from two sensors: an RGB-D camera on the one hand, and a radar on the other hand. We implement and evaluate two approaches for both the object localization and predicting the interception point. First, we present a baseline model that uses colour filtering for object localization in the RGB-D data. The interception point is then predicted using least squares regression over the ballistic trajectory equation. Second, we propose a deep learning based method that uses convolutional neural networks for both object localization and interception point prediction.

\begin{figure}
   \centering
   \includegraphics[width=1.4in]{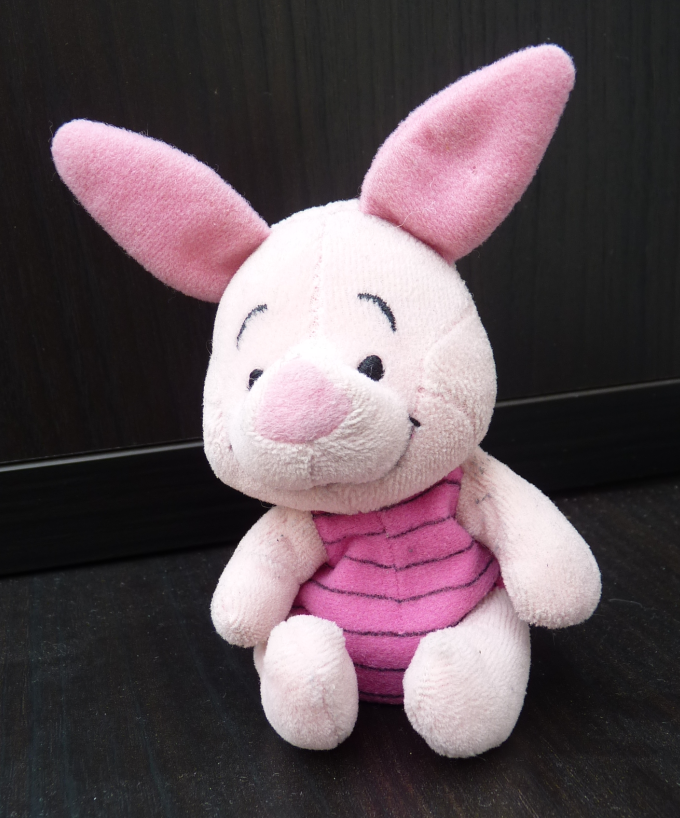}
   \caption{We throw a stuffed animal Piglet to the robotic manipulator.}
   \label{fig:piglet}
\end{figure}

In order to evaluate our approach, we perform actual throws towards a robotic arm. To ensure safety of everyone involved, we use a stuffed animal of Piglet\footnote{No piglets were harmed during these experiments} (Fig. \ref{fig:piglet}) in our experiments, which is also easy to detect in the RGB data for our baseline model. 

In the next section we will first discuss related work on object localization and trajectory prediction for catching objects in flight. In Section 3, we present our experimental hardware and software setup. Thereafter, we present our object localization and interception point prediction approaches, introducing both our baseline model and deep learning model. Finally we present our experimental results to end with our conclusions.

\section{RELATED WORK}
There are two prevailing techniques in literature on solving the problem of catching objects in flight: object localization and trajectory prediction.

\subsection{Object localization}

A common technique for obtaining the position of the object in an image is filtering out all pixels not matching the dominant colour of the target object~\cite{Mudjirahardjo}. Using the dominant colour of an object has the possibility to detect the location of arbitrarily shaped objects from the environment, as long as the dominant colours are not present in the background. Using the shape information of the objects offers a robust detection method against changing light conditions and similarly coloured objects in the scene. In 2D and 3D data, the two best performing solutions are Hough transforms and the RANSAC algorithm~\cite{Jacobs}. Using a shape based object localization method for detecting objects in point clouds is suggested by Heitz et al.~\cite{Heitz}. This is, however, only usable for detecting basic shapes, such as cube or ball shaped objects.

\subsection{Trajectory prediction}
Predicting the interception point between a thrown object and an end-effector is usually done using ballistic trajectory models or machine learning techniques.\newline

\subsubsection{Ballistic models}
Generally, the trajectory followed by a thrown object is a parabolic curve, only influenced by gravity and the air resistance of the object. 

Kumar et al. \cite{Kumar} use such a model to predict tennis ball trajectories during a tennis game. Due to the small size of the ball, they can neglect the impact of air resistance and only incorporate gravity in their predictions. They approximate the initial velocity and throwing angle based on the last two recorded points. Huang et al.~\cite{Huang} show that using all observed trajectory points yields improved results.
\newline
\subsubsection{Machine learning techniques}
The need for an exact physical model of motion is eliminated by Mironov and Pongratz~\cite{Mironov}. As they use nearest neighbour regression, they trade an explicit ballistic model for an increase in needed computations. One of the first successful deep learning approaches to catching objects in flight is presented by Ondruska and Posner~\cite{Ondruska}, with their unsupervised end-to-end mapping of raw sensor input to object trajectories by utilizing recurrent neural networks. 
Payeur and Gosselin~\cite{Payeur} achieve similar results by predicting position, orientation, velocity and acceleration of the object based on the coordinates of the object in the last 6 measurements.

\section{EXPERIMENTAL SETUP}
\subsection{Hardware overview}
\begin{figure}
    \centering
    \includegraphics[width=2.5in]{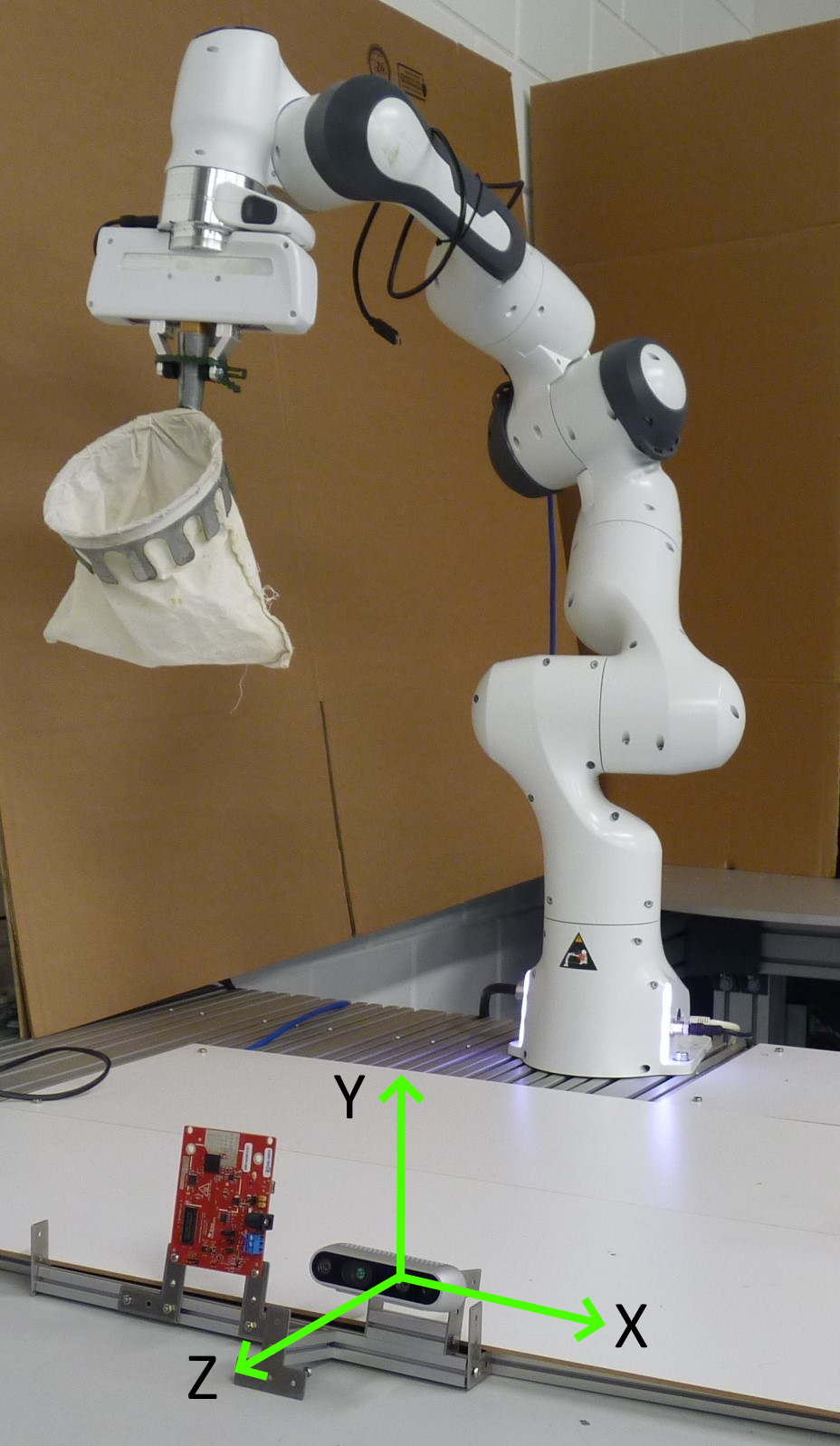}
    \caption{The experimental setup consisting of a Franka Emika robotic manipulator, an Intel Realsense RGB-D camera and a TI radar. Z is the depth axis, X the horizontal axis and Y the vertical axis}
    \label{fig:franka}
\end{figure}

An overview of our experimental setup and used coordinate system is shown in Fig. \ref{fig:franka}. We use a Franka Emika Panda robot~\cite{Emika} with a circular basket with a radius of 7.5cm mounted on the end-effector to intercept the thrown object. The sensors used for obtaining the location of the thrown object are the Intel RealSense D435 depth camera~\cite{Realsense} and TI mmWave IWR1443 radar~\cite{TI}.

\subsection{Software Architecture}
\begin{figure}
    \centering
    \includegraphics[width=\linewidth]{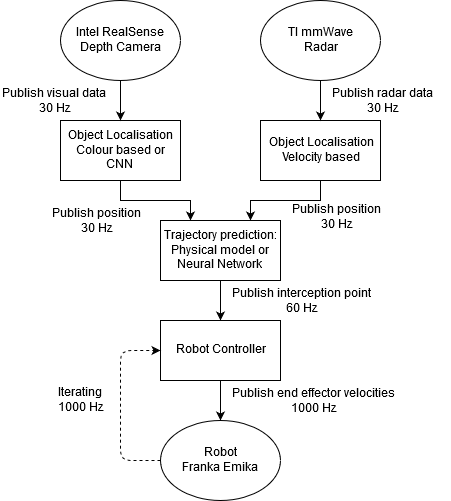}
    \caption{Block diagram of the application architecture. From both the RGBD camera and radar we extract position estimates of Piglet. These estimates are then fed to a trajectory prediction module to update the velocity controller of the robot.}
    \label{fig:arch}
\end{figure}

The application exists out of three main parts: object localization, trajectory prediction and controlling the robot movement. Each of the three parts is implemented as separate nodes and communicate using ROS~\cite{ros}. An overview of the high level system architecture is given in Fig. \ref{fig:arch}.

Moving the robot end-effector to the predicted interception point at a high velocity is crucial for completing the catching task. To move the robot, we adopt a velocity based controller that, based on the current state of the robot, calculates the distance on each axis between the current position and the goal. The controller iterates at a rate of 1000Hz and the goal is updated each time a new interception point is calculated.

\section{CATCHING APPROACH}
Catching an object consists of two separate stages: object localization and trajectory prediction. We consider two alternatives: a baseline approach and a deep learning approach.

\subsection{Baseline Approach}

Our baseline approach uses colour filtering to obtain a location estimate of the thrown object from the visual
sensor data. The Red-Green-Blue (RGB) data is converted into the Hue-Saturation-Value (HSV) colour space for easier colour comparison. The points outside the predefined colour range are removed. The remaining residue is converted into a single location by using the median over each axis. The TI mmWave radar provides Cartesian coordinates of moving targets. To filter out the thrown object from the background, we first remove all static background objects witch zero velocity, as well as all targets moving away from the radar. From the remaining values, the target with the highest velocity is kept for trajectory prediction.

To predict the interception point between the thrown object and the robot, we use a ballistic model. At any point in time, we fit a ballistic second degree curve to all available object localization points form both the RGBD and radar sensor, using first degree least squares regression. The resulting curve is then used to calculate the horizontal and vertical position of the interception point for a fixed depth value of -0.4m.

\subsection{Neural Network Approach}

In comparison to our baseline approach we also train two neural networks to perform the two stages. Using a data-driven neural network allows us to extend our approach to different objects and setups, as it is less prone to calibration (at the cost of data collection).
\newline
\subsubsection{Object localization}
\begin{figure}
    \centering
    \includegraphics[width=\linewidth]{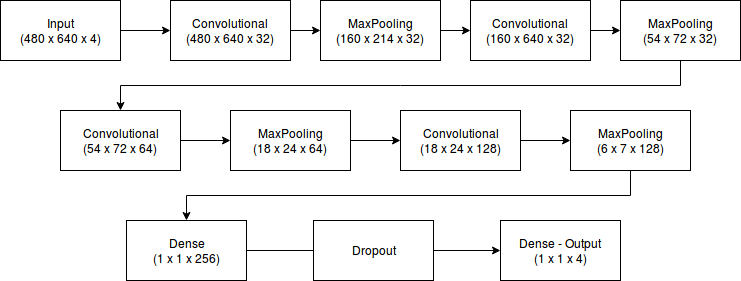}
    \caption{CNN architecture for detecting Piglet position from RGBD images}
    \label{fig:network}
\end{figure}

For object localization we use a convolutional neural network (CNN) to process the four channels (RGBD) captured by the depth camera at a resolution of 480 $\times$ 640 pixels. The CNN is based on the VGG16~\cite{Simonyan} architecture and consists out of four pairs of convolutional and maxpooling layers, followed by two dense layers, as shown in Fig.~\ref{fig:network}. We use the PReLU activation function after each layer.

The training data is collected using the experimental setup and labelled by the (x,y,z) coordinates of the baseline approach. The data is collected over multiple throws, under different environmental and lighting conditions. The resulting train set contains 5437 samples, with a separate test and validation set of 483 and 376 samples respectively. 
\newline
\subsubsection{Trajectory prediction}
\begin{figure}
    \centering
    \includegraphics[width=\linewidth]{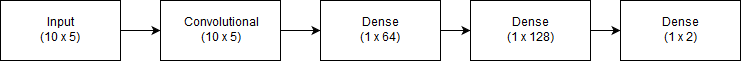}
    \caption{CNN architecture for predicting the interception point. The input consists of the last 10 detection points (or zeros if not available), containing the X,Y and Z coordinate as well as the time elapsed since the first point and which sensor provided the estimate. The output is the X and Y position of the interception point.}
    \label{fig:interception}
\end{figure}

To predict the interception point, we train again a CNN that takes the last ten localization points as input. For each localization point, we provide the (x,y,z) coordinate, as well as the time elapsed since the first point, and a 1 or 0 indicating the sensor that provided the estimate. When less than 10 points are available, we pad with zeros. Our CNN architecture is shown in  Fig. \ref{fig:interception}, and consists out of a single convolutional layer followed by three dense layers. We again use the PReLU activation function after each layer. The model provides two output values, namely the interception position on the horizontal and vertical axis.

We again collect a dataset using our experimental setup, where we throw Piglet and manually move the robot arm to the interception point. In case we moved too slowly and we failed to catch Piglet, we use the interception point calculated by the baseline approach. The resulting train set contains 379 unique trajectories while the test and validation set each contain 40 trajectories. To increase the amount of training data, we randomly shift the trajectories in the horizontal and vertical axis, increasing the amount of training data to 1895 samples.

We train both the object localization and interception point prediction CNN for 200 epochs with mean squared error (MSE) loss and Adam optimizer with learning rate of 1e-4.

\section{EXPERIMENTS}
We first evaluate the distance our robotic end-effector can move during throws of Piglet, after which we assess the overall catching performance.

\subsection{Time of Fligth}
To determine the available inference time for our model we experimentally determined the Time of Flight (TOF) of Piglet. We measured a minimal TOF between 815ms and 1292ms, a more detailed overview is given in Table~\ref{tab:TOF}. Note that the robotic end-effector can move approximately 60cm for the longer distance throw (6.70m), while only 20cm for the shorter distance throw (5.12m). Calculating a new interception point from sensor inputs takes between 128ms and 176ms, leaving enough time for moving the robot.

\begin{table}[b!]
    \centering
    \begin{tabular}{ c | c | c | c}
         Distance & Min. TOF & Max. TOF & Median TOF  \\ \hline
         5.12m    &  815ms   &  1062ms  & 974ms  \\
         6.70m    &  1135ms  &  1292ms  & 1206ms 
    \end{tabular}
    \caption{Time of Flight (TOF) measurements at different throwing distances}
    \label{tab:TOF}
\end{table}

\subsection{Catching Results}

\begin{figure}
    \centering
    \includegraphics[width=\linewidth]{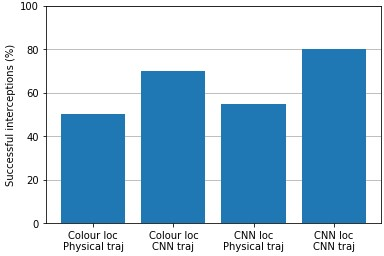}
    \caption{The overall catching performance of the different Piglet catching implementations}
    \label{fig:results}
\end{figure}

To evaluate the catching system, we record the success rate of catching Piglet for 20 throws, using any combination of the object localization and trajectory prediction approaches. The results are shown in Fig. \ref{fig:results}. The combination of a CNN trajectory prediction and the CNN object localization provides the most accurate results, with a catch ratio of 80\%. When using the colour filter object localization, the catch rate drops to 75\%. When using the physical trajectory prediction model, the catch ratio drops to about 50\%. We hypothesize that the lower performance of the physical trajectory prediction is due to the coarse approximations of the physical model, whereas the CNN can learn all relevant parameters influencing the trajectory, including small errors in the sensor calibration. We also noticed that the CNN object localization was more resilient to vertical motion blur compared to the colour filtering approach.

\section{Conclusions}
In this paper, we proposed a system for catching objects in flight, fusing information from both an RGBD sensor and a radar. We conducted experiments with a stuffed animal Piglet to validate our system. Our approach consists out of three parts: obtaining the location of the thrown object based on the camera input, predicting the object trajectory and moving the robot end-effector towards the interception point. We implemented and compared two methods for both object localization and interception point prediction: a colour filtering and ballistic curve fitting baseline as well as deep learning method using CNNs. We show that we are able to process in real-time in order to move the robotic arm in time for a successful catch. Moreover, our CNN based approach achieves a success rate of 80\%. 

As future work we will further research to what extend we can generalize the CNN approach to catch different objects with multiple shapes and appearances. Also, we would like to investigate a single object detection and interception point prediction pipeline that is trained end-to-end, instead of two separate steps.

\section*{Acknowledgments}
Ozan \c{C}atal is funded by a Ph.D. grant of the Flanders Research  Foundation (FWO).

\addtolength{\textheight}{-12cm}   






\begin{thebibliography}{99}
\bibitem{Mudjirahardjo} P. Mudjirahardjo, “Color feature based object localization in real
time implementation,” IOSR Journal of Computer Engineering (IOSR-JCE), vol. 20, no. 2, pp. 31 – 37, 2018. [Online]. Available: http://www.iosrjournals.org/iosr-jce/papers/Vol20-issue2/Version-
3/F2002033137.pdf
\bibitem{Jacobs} L. Jacobs, J. Weiss, and D. Dolan, “Object tracking in noisy radar data:
Comparison of hough transform and ransac,” in IEEE International Con-
ference on Electro-Information Technology , EIT 2013, May 2013, pp. 1–6.
\bibitem{Heitz} G. Heitz, G. Elidan, B. Packer, and D. Koller, “Shape-based object
localization for descriptive classification,” in Advances in Neural
Information Processing Systems 21, D. Koller, D. Schuurmans, Y. Bengio,
and L. Bottou, Eds. Curran Associates, Inc., 2009, pp. 633–640.
[Online]. Available: http://papers.nips.cc/paper/3540-shape-based-object-
localization-for-descriptive-classification.pdf
\bibitem{Kumar} A. Kumar, P. S. Chavan, V. K. Sharatchandra, S. David, P. Kelly, and N. E.
O’Connor, “3d estimation and visualization of motion in a multicamera
network for sports,” in 2011 Irish Machine Vision and Image Processing
Conference, Sep. 2011, pp. 15–19.
\bibitem{Huang} C. Huang, W.-J. Tsai, S.-Y. Lee, and J.-Y. Yu, “Ball tracking and 3d tra-
jectory approximation with applications to tactics analysis from single-
camera volleyball sequences,” Multimedia Tools and Applications - MTA,
vol. 60, 10 2012.
\bibitem{Mironov} K. Mironov and M. Pongratz, “Fast knn-based prediction for the trajectory
of a thrown body,” in 2016 24th Mediterranean Conference on Control and
Automation (MED), June 2016, pp. 512–517.
\bibitem{Ondruska} P. Ondruska and I. Posner, “Deep tracking: Seeing beyond seeing using
recurrent neural networks,” CoRR, vol. abs/1602.00991, 2016. [Online].
Available: http://arxiv.org/abs/1602.00991
\bibitem{Payeur}
P. Payeur, , and C. M. Gosselin, “Trajectory prediction for moving objects
using artificial neural networks,” IEEE Transactions on Industrial Elec-
tronics, vol. 42, no. 2, pp. 147–158, April 1995.
\bibitem{Emika}
F. Emika, “Franka emika panda,” 2019. [Online]. Available:
https://www.franka.de/panda
\bibitem{Realsense}Intel, “Intel realsense,” 2019. [Online]. Available: https://realsense.intel.com/
\bibitem{TI} T. Instruments, “Ti mmwave iwr1443,” 2019. [Online]. Available:
http://www.ti.com/product/IWR1443
\bibitem{ros} O. S. R. Foundation, “Robot operating system,” 2019. [Online]. Available:
http://www.ros.org/
\bibitem{Simonyan}
K. Simonyan and A. Zisserman, “Very deep convolutional networks for
large-scale image recognition,” arXiv 1409.1556, 09 2014.



\end{thebibliography}
\end{document}